\PassOptionsToPackage{dvipsnames,table}{xcolor}
\documentclass[11pt]{article}
\usepackage{acl}

\usepackage{times}
\usepackage{latexsym}
\usepackage{booktabs}
\usepackage{amsmath,amssymb}
\usepackage{enumitem}
\usepackage{subcaption}
\usepackage{multirow}
\usepackage{graphicx}
\usepackage{xspace}

\usepackage[T1]{fontenc}

\usepackage[utf8]{inputenc}

\usepackage{microtype}

\usepackage{inconsolata}

\newcommand{\fig}{Fig.\xspace}

\usepackage[most]{tcolorbox}

\newtcolorbox{interfacebox}[1]{
  colback=gray!3,
  colframe=gray!55,
  title=\textbf{#1},
  fonttitle=\small,
  fontupper=\footnotesize,
  boxrule=0.4pt,
  arc=1pt,
  left=4pt,
  right=4pt,
  top=4pt,
  bottom=4pt,
  breakable
}

\newtcolorbox{promptbox}[1]{
  colback=gray!2,
  colframe=black!55,
  title=\textbf{#1},
  fonttitle=\small,
  fontupper=\footnotesize,
  boxrule=0.5pt,
  arc=1pt,
  left=4pt,
  right=4pt,
  top=4pt,
  bottom=4pt,
  breakable
}

\newtcblisting{promptlisting}[1]{
  breakable,
  colback=gray!2,
  colframe=gray!35,
  coltitle=black,
  colbacktitle=gray!10,
  title=\textbf{#1},
  fonttitle=\footnotesize,
  boxrule=0.35pt,
  arc=1pt,
  left=3pt,
  right=3pt,
  top=3pt,
  bottom=3pt,
  listing only,
  listing options={
    basicstyle=\scriptsize\ttfamily,
    breaklines=true,
    breakatwhitespace=false,
    columns=fullflexible,
    keepspaces=true,
    showstringspaces=false
  }
}

\newtcolorbox{compactbox}[1]{
  breakable,
  colback=gray!2,
  colframe=gray!35,
  coltitle=black,
  colbacktitle=gray!10,
  title=\textbf{#1},
  fonttitle=\footnotesize,
  fontupper=\footnotesize,
  boxrule=0.35pt,
  arc=1pt,
  left=4pt,
  right=4pt,
  top=4pt,
  bottom=4pt
}

\title{CityPlanner: A Sandbox Agent for Executable Urban Planning}

\author{
  Wentao Zhang, Jingyuan Wang$^{\dagger}$, Zetong Zhou,Yifan Yang, Wenrui Wang \\
  School of Computer Science and Engineering, Beihang University, Beijing, China \\
  MIIT Key Laboratory of Data and Decision Intelligence, Beihang University, Beijing, China \\
  \texttt{\symbol{123}zhangwt97\symbol{125}@buaa.edu.cn}
}

\begin{document}
\maketitle
\begin{abstract}
Urban planning is a real-world spatial optimization problem that requires selecting feasible actions from large candidate spaces under practical objectives such as cost and service quality. Existing optimization and reinforcement learning methods are effective for fixed formulations, but often depend on task-specific representations and constraint handling. We propose \emph{CityPlanner}, a sandbox-agent framework for executable urban planning. CityPlanner introduces \emph{UrbanSandbox}, a unified file-based environment where agents inspect task files, generate plans, run evaluators, and revise decisions based on executable feedback. To make learning tractable, we further propose atomic-task reinforcement learning, which decomposes long sandbox trajectories into \emph{BuildPlan} for initial construction and \emph{ImprovePlan} for feedback-based refinement. Experiments on a real-world benchmark show that CityPlanner consistently outperforms heuristic, task-specific RL, and general LLM-agent baselines. Ablations verify the contributions of UrbanSandbox, atomic-task RL, and iterative deployment. We release the code and dataset at \url{https://anonymous.4open.science/r/co-agent-C1C8}.

\end{abstract}

\section{Introduction}

Urban planning is a fundamental problem in city management. It converts spatial data, infrastructure conditions, and policy objectives into concrete decisions over land use, road construction, and public facilities~\citep{zheng2025urban,liu2025large,zheng2023road,von2022reinforcement,liu2023optimal}. These decisions directly affect transportation efficiency, service accessibility, and ecological quality. From an optimization perspective, urban planning is challenging because it involves large combinatorial action spaces, hard feasibility constraints, and multiple competing objectives, such as construction cost and environmental benefit. The value of each planning action is also highly context-dependent, as it depends on surrounding population, topology, and local demand patterns.

\begin{figure}[tbp]
    \centering
    \includegraphics[width=\linewidth]{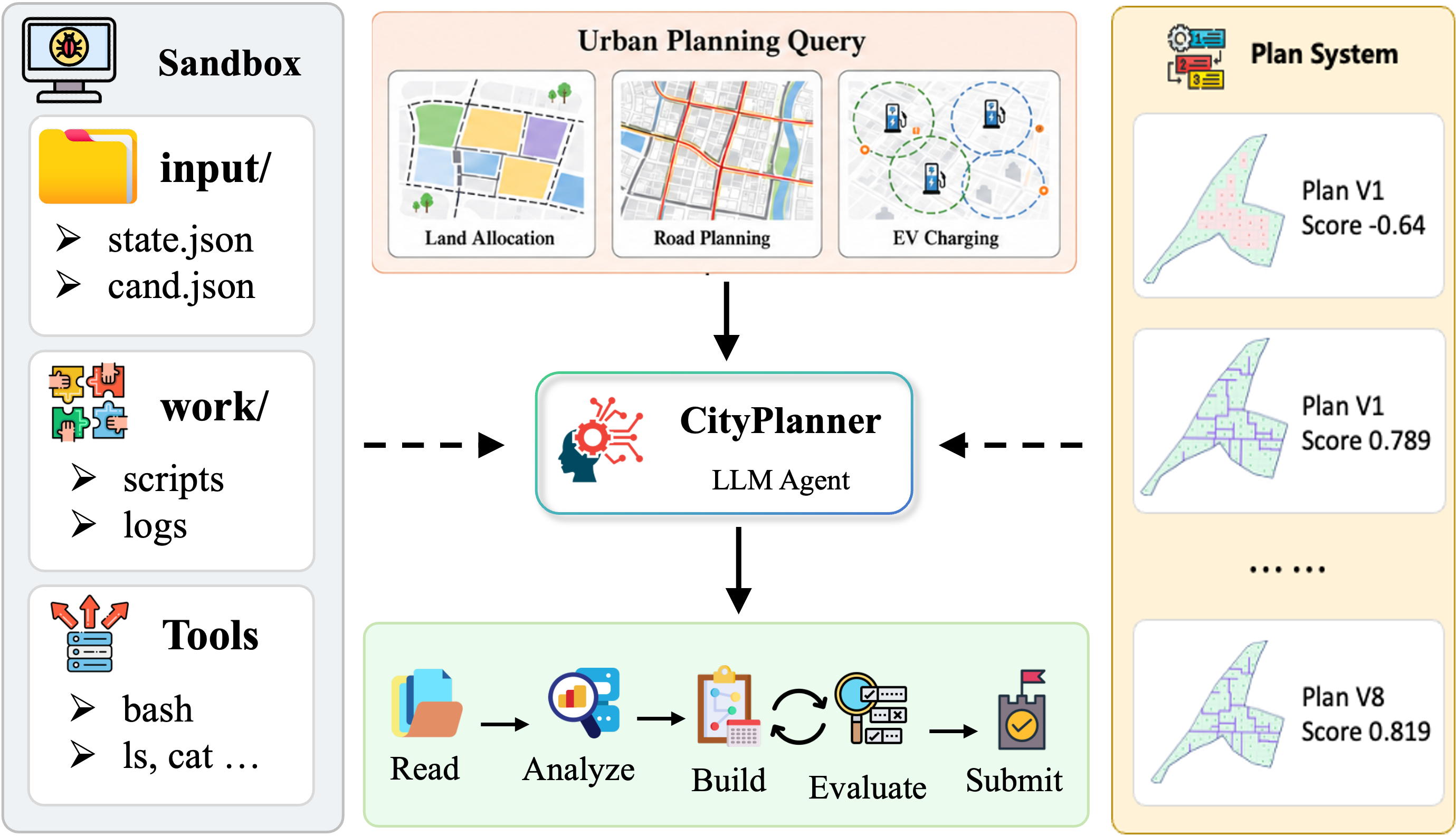}
    \caption{Executable urban planning with CityPlanner.}
    \label{fig:demo}
\end{figure}

A large body of work has studied urban planning with optimization, heuristic search, and reinforcement learning methods~\citep{rasheed2020deep,von2022reinforcement,zheng2023road}. These methods are effective when the state representation, action space, constraints, and objective function are fixed in advance. However, practical urban planning tasks are heterogeneous: different tasks may involve different candidate types, constraint definitions, evaluation procedures, and feedback forms. Methods designed for one task often require non-trivial redesign when transferred to another scenario, limiting their applicability.

Large language model agents provide a possible interface for such heterogeneous planning tasks. Instead of relying on a predefined vectorized state-action representation, an LLM agent can read task descriptions, inspect structured files, write intermediate analyses, execute tools, and revise solutions according to feedback~\citep{yao2023react,shinn2023reflexion,qintoolllm,patilgorilla}. Nevertheless, directly using an LLM as a black-box planner is insufficient for constrained urban planning. A generated plan must satisfy explicit feasibility constraints and be evaluated by task-specific programs rather than natural-language plausibility. The key question is how to leverage the broad capabilities of LLM to effectively solve urban planning problems.

In this paper, we formulate urban planning as an executable sandbox-agent problem. The agent is placed in a sandbox containing city data, candidate actions, task constraints, executable evaluators, and output interfaces. As shown in \fig~\ref{fig:demo}, the agent solves the task by reading input files, constructing a plan, running the evaluator, observing diagnostic feedback, and revising the plan. This formulation provides a common interaction protocol for heterogeneous urban planning tasks while preserving their task-specific objectives and constraints.

Training an LLM agent in such a sandbox remains non-trivial. Complete sandbox trajectories are long and noisy, containing repeated script execution, invalid submissions, and intermediate revisions. As the interaction proceeds, outdated observations and failed attempts accumulate in the context. Moreover, the main planning reward is usually available only after a complete plan is submitted and evaluated, leading to sparse rewards and difficult long-horizon credit assignment. Direct reinforcement learning over complete trajectories can therefore improve superficial executable behaviors while still failing to produce high-quality plans.

To address these issues, we propose \emph{CityPlanner}, a sandbox-agent framework for executable urban planning. CityPlanner has two components. First, \emph{UrbanSandbox} provides a unified file-based environment for heterogeneous planning tasks. The agent interacts with the environment through a minimal \texttt{bash}-based interface, enabling file inspection, plan generation, script execution, and evaluator invocation without task-specific agent APIs. Second, \emph{atomic-task reinforcement learning} decomposes long-horizon sandbox planning into two finite-context tasks: \emph{BuildPlan}, which constructs an initial feasible plan, and \emph{ImprovePlan}, which revises an existing plan using evaluator diagnostics. At inference time, CityPlanner composes the two skills into an iterative procedure that repeatedly evaluates and improves the best-so-far solution.

For empirical evaluation, we collect real-world data from OpenStreetMap~\cite{OpenStreetMap} and build an executable benchmark covering 5,670 planning instances across three tasks. Experiments show that CityPlanner consistently improves planning quality over heuristic, task-specific RL, and LLM-agent baselines, while ablations confirm the contributions of UrbanSandbox, atomic-task RL, and iterative refinement.

\section{Related Work}


\paragraph{LLM agents and executable environments.}
Recent LLM agents extend language models to interactive problem solving with tools, APIs, files, and executable environments~\citep{yao2023react,shinn2023reflexion,qintoolllm,patilgorilla}. Existing agent benchmarks mainly focus on general tool use, web interaction, coding, or command-line tasks~\citep{vishwakarma2025can,pysklo2026agent,merrill2026terminal,openthoughts-agent}. In parallel, LLMs have been explored for urban analysis, planning assistance, and decision support~\citep{jiang2024urbanllm,zhu2024plangpt,zhou2402large,zheng2025urban,agrawal2025large,liu2025large}. Different from language-only planning assistance, CityPlanner studies executable urban planning, where the agent must submit structured spatial plans that are verified and scored by task-specific evaluators.

\paragraph{Reinforcement learning for LLM agents.}
Reinforcement learning has been used to improve LLM agents in tool-use, web navigation, coding, and interactive planning environments~\citep{li2025encouraging,qiwebrl,cheng2025agent,feng2025retool,qian2026toolrl,tan2025process}. A common difficulty is that agent trajectories are long, noisy, and often rewarded only after task completion. CityPlanner addresses this issue by decomposing executable urban planning into two finite-context atomic tasks, enabling construction and refinement to be trained with separate task-level rewards.

\section{Problem Statement}
\label{subsec:problem}

Urban planning aims to improve an existing urban region by selecting planning actions under task-specific constraints. We consider following tasks:

\noindent~$\bullet$ \emph{Land allocation} assigns land-use types to candidate parcels under land-use, spatial, ecological, and service-balance constraints.

\noindent~$\bullet$ \emph{Road construction} selects road segments to add to the existing network under budget, quantity, and spatial-validity constraints.

\noindent~$\bullet$ \emph{Station placement} selects candidate sites for new charging facilities under demand, capacity, infrastructure, and cost constraints.

Although these tasks have different spatial objectives, they share the same decision structure: given an urban region and a finite set of candidate actions, the planner selects a constraint-satisfying subset as the final plan.
Formally, each instance $p$ is represented by an attributed city graph
\begin{equation}
\mathcal{G}_p=(V_p,E_p,\mathbf{X}^V_p,\mathbf{X}^E_p),
\end{equation}
where $V_p$ denotes urban spatial units, $E_p$ denotes road-network edges, and $\mathbf{X}^V_p,\mathbf{X}^E_p$ denote their attributes. Each instance provides a candidate set $\mathcal{C}_p\subseteq V_p\cup E_p$. A plan $P\subseteq\mathcal{C}_p$ selects a subset of candidates. We use $\Psi_p(P,\mathcal{G}_p)\in\{0,1\}$ to denote whether $P$ satisfies all task-specific constraints, and $S_p(P)$ to denote its planning score. The goal is
\begin{equation}
P_p^\star
=
\arg\max_{P\subseteq\mathcal{C}_p:\Psi_p(P,\mathcal{G}_p)=1}
S_p(P).
\end{equation}
Detailed definitions are provided in Appendix~\ref{app:task_definitions}.

\section{Method}
\label{sec:method}
As shown in Figure~\ref{fig:demo}, CityPlanner solves these tasks through UrbanSandbox, which provides executable plan generation, evaluation, and feedback. It further decomposes long planning trajectories into two atomic tasks, and composes them iteratively at inference time for long-horizon planning.

\subsection{UrbanSandbox}
\label{subsec:urbansandbox}

The problem statement defines the planning objective. UrbanSandbox specifies how each planning instance is exposed to an agent as an executable environment. For each instance $p$, we construct
\begin{equation}
\mathcal{E}_p =
\langle
\mathcal{I}_p,\mathcal{W}_p,\mathcal{O}_p,
\operatorname{Exec}_p,\operatorname{Eval}_p
\rangle .
\end{equation}
Here, $\mathcal{I}_p$ is the read-only input space, $\mathcal{W}_p$ is the writable workspace, $\mathcal{O}_p$ is the output space, $\operatorname{Exec}_p$ executes commands, and $\operatorname{Eval}_p$ evaluates plans.

UrbanSandbox is implemented as a file-based sandbox. The \texttt{input/} directory contains the visible city state, candidate set, task schema, constraints, and evaluator. The \texttt{work/} directory is used for intermediate analyses and scripts. The \texttt{outputs/} directory stores submitted plans and evaluation results. Detailed definitions are provided in Appendix~\ref{app:sandbox_prompt}.

The agent interacts with UrbanSandbox through a minimal \texttt{bash}-based interface. At step $t$, the policy emits a shell command
\begin{equation}
a_t\sim\pi_\theta(\cdot\mid h_t),
\end{equation}
where $h_t=(o_0,a_0,\ldots,o_t)$ denotes the interaction history. The sandbox executes the command and returns an observation:
\begin{equation}
o_{t+1}=\operatorname{Exec}_p(a_t).
\end{equation}
The observation may contain file contents, standard output, execution failures, parsing errors, or evaluator diagnostics. The trajectory is written as
\begin{equation}
\tau=(o_0,a_0,o_1,a_1,\ldots,o_T).
\end{equation}

The final plan is parsed from \texttt{outputs/final\_plan.json}. UrbanSandbox then checks whether the plan satisfies the aggregate constraint function $\Psi_p(P,\mathcal{G}_p)$ and evaluates its planning score:
\begin{equation}\small
\operatorname{Eval}_p(P)=
\begin{cases}
S_p(P), & \Psi_p(P,\mathcal{G}_p)=1,\\
\operatorname{Error}_p(P), & \Psi_p(P,\mathcal{G}_p)=0.
\end{cases}
\end{equation}
Thus, plan validity and quality are determined by executable verification rather than language check. 

\begin{figure}[tbp]
    \centering
    \includegraphics[width=\linewidth]{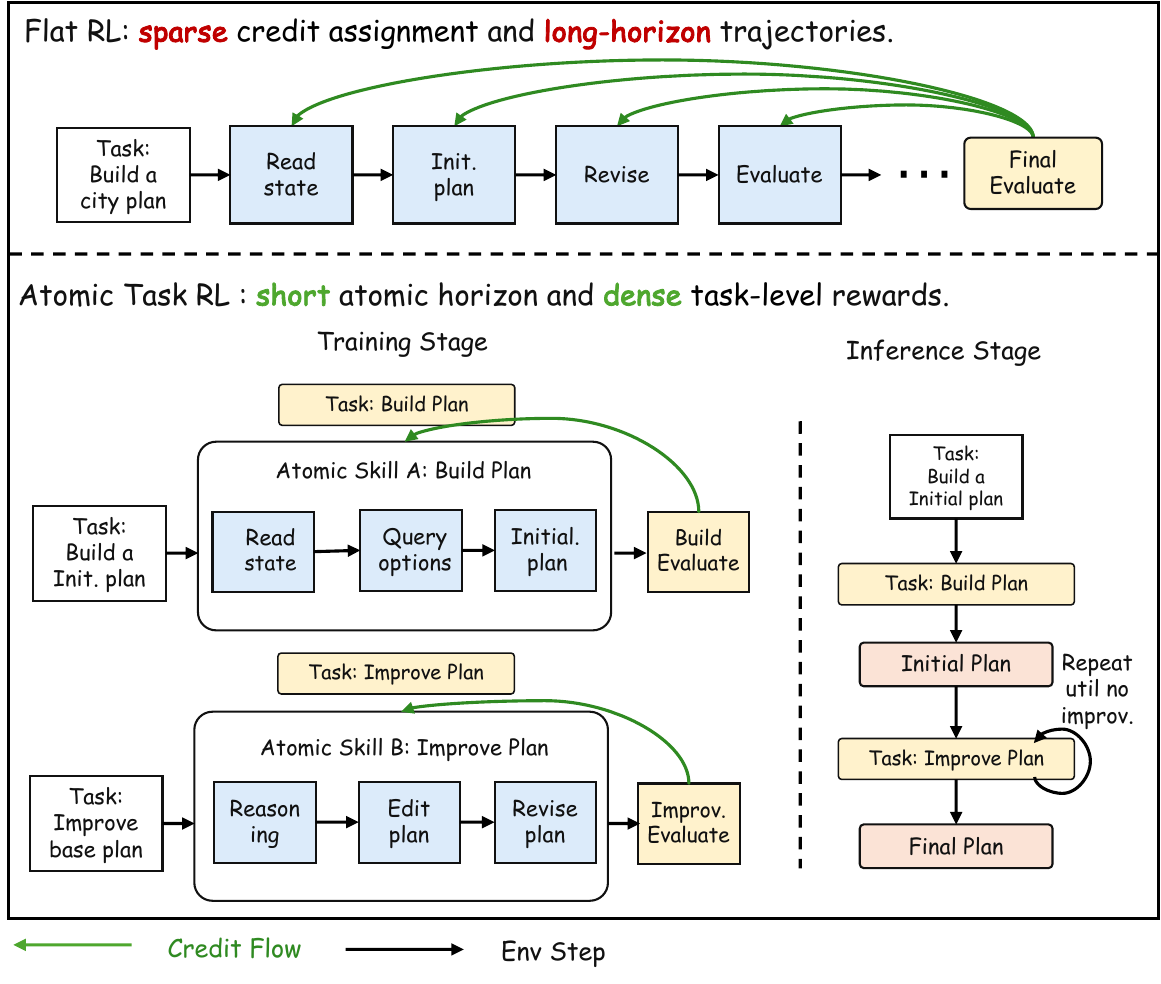}
    \caption{Atomic-task reinforcement learning. CityPlanner decomposes long-horizon sandbox planning into BuildPlan and ImprovePlan, and composes the two learned skills into iterative refinement at inference time.}
    \label{fig:atr}
\end{figure}

\subsection{Atomic-Task Reinforcement Learning}
\label{subsec:atomic_learning}
A pretrained language model provides useful priors for file inspection, structured generation, script writing, and tool use, but it is not by itself a reliable planner under hard spatial constraints. As shown in Figure~\ref{fig:atr}, directly training on complete sandbox trajectories requires assigning a sparse final reward to many preceding steps, such as state reading, plan construction, revision, and evaluation. These long and noisy trajectories make credit assignment difficult. CityPlanner therefore decomposes training into two atomic tasks, following the common structure of strong heuristic solvers: first construct an initial solution and then improve it by search. BuildPlan learns the construction stage from task files, while ImprovePlan learns the improvement stage from the incumbent plan and evaluator diagnostics. Each task has a shorter horizon and an independent task-level reward.

We use $\operatorname{Rollout}_{\pi_\theta}^{z}(\cdot)$ to denote a bounded multi-turn interaction between the policy and UrbanSandbox under atomic task $z$, whose output is the final plan submitted to \texttt{outputs/final\_plan.json}.

\paragraph{BuildPlan.}
BuildPlan is the construction task. Given the sandbox environment $\mathcal{E}_p$, which contains the city graph $\mathcal{G}_p$, candidate set $\mathcal{C}_p$, task constraints, and output interface, the policy generates an initial plan through a bounded sandbox rollout:
\begin{equation}
P_0=\operatorname{Rollout}_{\pi_\theta}^{\mathrm{build}}(\mathcal{E}_p).
\end{equation}
This task trains the model to construct a feasible executable plan from scratch.

\paragraph{ImprovePlan.}
ImprovePlan is the revision task. Given the sandbox environment $\mathcal{E}_p$, the current plan $P_t$, and evaluator diagnostics $\mathbf{d}_t$, the policy generates a revised plan:
\begin{equation}
\tilde{P}_{t+1}
=
\operatorname{Rollout}_{\pi_\theta}^{\mathrm{imp}}
(\mathcal{E}_p,P_t,\mathbf{d}_t),
\end{equation}
where $\mathbf{d}_t$ denotes the constraint and score feedback returned by UrbanSandbox. This task trains the model to improve an existing plan from feedback.

\paragraph{Reward.}
Both atomic tasks use the same plan-level terminal reward. Given a submitted plan $P$, the reward is its planning score if it satisfies the task constraints, and a failure penalty otherwise:
\begin{equation}
R_p(P)=
\begin{cases}
S_p(P), & \Psi_p(P,\mathcal{G}_p)=1,\\
R_{\mathrm{fail}}, & \text{otherwise}.
\end{cases}
\end{equation}

\paragraph{Learning Algorithm.}
We optimize the policy with GRPO~\cite{grpo} by sampling rollouts from BuildPlan and ImprovePlan. Each rollout receives the corresponding task-level reward. This trains the policy on bounded atomic skills, avoiding direct optimization over long and noisy trajectories.

\subsection{Iterative Inference}
\label{subsec:inference}
At inference time, CityPlanner composes the two learned atomic skills into an iterative planning procedure, as shown in Figure~\ref{fig:atr}. Given a test environment $\mathcal{E}_p$, CityPlanner first invokes BuildPlan to generate an initial plan $P_0$. UrbanSandbox evaluates this plan and returns its score and diagnostic feedback $\mathbf{d}_0$.
CityPlanner then repeatedly invokes ImprovePlan. At iteration $t$, the current best plan $P_t$ and diagnostics $\mathbf{d}_t$ are provided through the sandbox, and the model proposes a revised plan $\tilde{P}_{t+1}$. The proposal is then evaluated by UrbanSandbox before being accepted or rejected.

After each ImprovePlan call, UrbanSandbox evaluates the proposed plan. We accept a proposal only if it is better under a validity-first order: feasible plans are preferred to infeasible ones, and among feasible plans, higher scores are preferred. Denote this order by $\succ_p$. The update rule is
\begin{equation}
P_{t+1}=
\begin{cases}
\tilde{P}_{t+1}, & \tilde{P}_{t+1}\succ_p P_t,\\
P_t, & \text{otherwise}.
\end{cases}
\end{equation}
The loop stops after a fixed iteration budget or a patience budget, and returns the best-so-far plan.

\begin{table}[t]
\centering
\caption{Statistics of the OSM-based benchmark.}
\label{tab:data_stats}
\scriptsize
\setlength{\tabcolsep}{3.2pt}
\begin{tabular}{l c ccc ccc}
\toprule
\multirow{2}{*}{\textbf{Task}}
& \multirow{2}{*}{\textbf{Train/Test}}
& \multicolumn{3}{c}{\textbf{\#Instances}}
& \multicolumn{3}{c}{\textbf{Avg. \#Candidates}} \\
\cmidrule(lr){3-5}
\cmidrule(lr){6-8}
&
& \textbf{Small} & \textbf{Medium} & \textbf{Large}
& \textbf{Small} & \textbf{Medium} & \textbf{Large} \\
\midrule
LA
& 1,512 / 378 & 816 & 745 & 329 & 85 & 119 & 199 \\
RC
& 1,512 / 378 & 708 & 780 & 402 & 63 & 102 & 143 \\
SP
& 1,512 / 378 & 816 & 745 & 329 & 80 & 80 & 180 \\
\midrule
Total
& 4,536 / 1,134 & 2,340 & 2,270 & 1060 & -- & -- & -- \\
\bottomrule
\end{tabular}
\end{table}

\section{Experiments}
\label{sec:exp}



\begin{table*}[t]
\centering
\caption{Performance comparison across tasks and difficulty levels. Results are reported as mean$\pm$std. The best result in each column is highlighted in bold, and the second-best result is underlined.}
\label{tab:overall_difficulty_std}
\scriptsize
\setlength{\tabcolsep}{2.5pt}
\resizebox{\textwidth}{!}{
\begin{tabular}{lccccccccc}
\toprule
\multirow{2}{*}{\textbf{Method}}
& \multicolumn{3}{c}{\textbf{Land Allocation}}
& \multicolumn{3}{c}{\textbf{Road Construction}}
& \multicolumn{3}{c}{\textbf{Station Placement}} \\
\cmidrule(lr){2-4}
\cmidrule(lr){5-7}
\cmidrule(lr){8-10}
& \textbf{Small} & \textbf{Medium} & \textbf{Large}
& \textbf{Small} & \textbf{Medium} & \textbf{Large}
& \textbf{Small} & \textbf{Medium} & \textbf{Large} \\
\midrule

\multicolumn{10}{l}{\textit{\textbf{Heuristic Methods}}} \\
\addlinespace[1pt]
GRASP
& 0.391$\pm$0.042
& 0.448$\pm$0.058
& 0.441$\pm$0.034
& \textbf{0.717$\pm$0.011}
& \underline{0.729$\pm$0.019}
& \underline{0.690$\pm$0.018}
& 0.279$\pm$0.024
& 0.064$\pm$0.009
& 0.020$\pm$0.001 \\

SA
& 0.395$\pm$0.086
& 0.441$\pm$0.097
& 0.431$\pm$0.044
& \underline{0.662$\pm$0.056}
& 0.567$\pm$0.065
& 0.596$\pm$0.074
& \underline{0.610$\pm$0.052}
& 0.437$\pm$0.094
& 0.353$\pm$0.078 \\

ALNS
& \underline{0.396$\pm$0.091}
& \underline{0.456$\pm$0.096}
& \underline{0.451$\pm$0.073}
& 0.653$\pm$0.063
& 0.513$\pm$0.059
& 0.510$\pm$0.077
& 0.501$\pm$0.073
& 0.434$\pm$0.080
& \underline{0.392$\pm$0.075} \\

\midrule
\multicolumn{10}{l}{\textit{\textbf{TSRL Methods}}} \\
\addlinespace[1pt]
MLP+PPO
& 0.205$\pm$0.015
& 0.276$\pm$0.011
& 0.271$\pm$0.014
& 0.070$\pm$0.066
& 0.166$\pm$0.069
& 0.135$\pm$0.049
& 0.222$\pm$0.021
& 0.132$\pm$0.011
& 0.153$\pm$0.018 \\

AM+PPO
& 0.184$\pm$0.014
& 0.248$\pm$0.011
& 0.248$\pm$0.010
& 0.269$\pm$0.032
& 0.256$\pm$0.025
& 0.307$\pm$0.010
& 0.224$\pm$0.017
& 0.175$\pm$0.026
& 0.194$\pm$0.011 \\

GNN+PPO
& 0.212$\pm$0.016
& 0.289$\pm$0.010
& 0.280$\pm$0.012
& 0.174$\pm$0.064
& 0.147$\pm$0.069
& 0.168$\pm$0.045
& 0.185$\pm$0.017
& 0.121$\pm$0.049
& 0.118$\pm$0.055 \\

\midrule
\multicolumn{10}{l}{\textit{\textbf{LLM Agents}}} \\
\addlinespace[1pt]
Qwen3.5
& 0.366$\pm$0.061
& 0.329$\pm$0.093
& 0.370$\pm$0.090
& 0.332$\pm$0.049
& 0.498$\pm$0.044
& 0.478$\pm$0.045
& 0.503$\pm$0.032
& 0.396$\pm$0.046
& 0.379$\pm$0.061 \\

Mimo V2.5
& 0.335$\pm$0.072
& 0.317$\pm$0.081
& 0.344$\pm$0.096
& 0.414$\pm$0.042
& 0.485$\pm$0.037
& 0.419$\pm$0.015
& 0.415$\pm$0.021
& 0.339$\pm$0.058
& 0.326$\pm$0.073 \\

DS-V4
& 0.323$\pm$0.095
& 0.364$\pm$0.083
& 0.360$\pm$0.056
& 0.334$\pm$0.035
& 0.579$\pm$0.038
& 0.527$\pm$0.039
& 0.527$\pm$0.021
& \underline{0.444$\pm$0.058}
& 0.381$\pm$0.052 \\

CityPlanner
& \textbf{0.402$\pm$0.014}
& \textbf{0.459$\pm$0.031}
& \textbf{0.451$\pm$0.026}
& 0.609$\pm$0.021
& \textbf{0.735$\pm$0.014}
& \textbf{0.704$\pm$0.054}
& \textbf{0.612$\pm$0.060}
& \textbf{0.462$\pm$0.034}
& \textbf{0.397$\pm$0.062} \\

\bottomrule
\end{tabular}
}
\end{table*}

\subsection{Experimental Setup}
\label{subsec:exp_setup}
\paragraph{Benchmark.}
We construct the benchmark from large-scale OpenStreetMap data over diverse urban regions in China, with details of data collection and preprocessing provided in Appendix~\ref{app:dataset_details}. The benchmark contains 5,670 planning instances over 1,890 urban tiles across three tasks. 

\paragraph{Baselines.}
We compare CityPlanner with three groups of baselines: \textbf{heuristic methods}, including GRASP~\cite{feo1995greedy}, SA~\cite{van1987simulated}, and ALNS~\cite{ahuja2000very}; \textbf{task-specific reinforcement learning methods}, including MLP+PPO, AM+PPO, and GNN+PPO; and \textbf{general LLM agents}, including Qwen3.5~\cite{qwen35}, Mimo V2.5~\cite{mimov25}, and DS-V4~\cite{deepseekv4}, which solve the tasks through UrbanSandbox without task-specific training. CityPlanner is based on Qwen3-8B and is post-trained in two stages: we first adapt the model to a Terminal-Bench-style interaction format using the public OT-SFT data~\cite{ot-agent}, and then train it with atomic-task RL on urban tasks. Implementation details of all methods are provided in Appendix~\ref{app:implementation}.

\paragraph{Evaluation protocol.}
All methods are evaluated on the same instances. Agent-based methods use the same sandbox files, output format, and interaction budget. In the main comparison, we use the task-specific objective score returned by the evaluator as the primary metric. The score definitions follow the original task objectives and are detailed in Appendix~\ref{app:task_definitions}. For all tasks, higher scores indicate better planning quality. Each method is evaluated three times with three different random seeds.

\subsection{Overall Performance}
\label{subsec:overall_performance}
Table~\ref{tab:overall_difficulty_std} reports the main comparison across three tasks and three difficulty levels. We summarize the results as follows.

\noindent~$\bullet$ \emph{CityPlanner achieves the strongest overall performance.}
CityPlanner obtains the best score in 8 out of 9 settings. This indicates that CityPlanner is consistently effective across heterogeneous planning tasks and scales.

\noindent~$\bullet$ \emph{Strong heuristics remain competitive baselines.}
Among non-agent baselines, heuristic methods generally perform best. For example, GRASP achieves the best score on small Road Construction and remains the second-best method on medium and large Road Construction. SA is also highly competitive on small Station Placement. These results show that search heuristics are still strong, especially on smaller or more regular instances, but they require task-specific designs for search moves, repair rules, and constraint handling.

\noindent~$\bullet$ \emph{Task-specific RL methods are less effective.}
The TSRL baselines perform substantially worse across all tasks. 
This suggests that, under limited training data, neural policies fail to learn effective planning strategies for these urban planning tasks.

\noindent~$\bullet$ \emph{Agent-based methods are competitive on complex settings.}
General LLM agents achieve competitive results in several larger settings, suggesting that tool use and iterative reasoning are useful for complex planning tasks. CityPlanner further strengthens this trend by combining executable feedback with atomic-task training and iterative refinement. These results indicate that agent-based planning becomes more effective as the candidate space grows and direct search becomes harder.

\label{subsec:scaffold_effect}

\begin{table}[t]
\centering
\caption{Comparison of test-time planning strategies using DeepSeek as a frozen planner. 
Ours (Scaffold-only) denotes our UrbanSandbox-based atomic workflow without post-training.}
\label{tab:deepseek_scaffold}
\resizebox{\columnwidth}{!}{
\begin{tabular}{lcccccc}
\toprule
\multirow{2}{*}{\textbf{Method}}
& \multicolumn{2}{c}{\textbf{Land Allocation}}
& \multicolumn{2}{c}{\textbf{Road Construction}}
& \multicolumn{2}{c}{\textbf{Station Placement}} \\
\cmidrule(lr){2-3}
\cmidrule(lr){4-5}
\cmidrule(lr){6-7}
& \textbf{Feas.}$\uparrow$ & \textbf{Score}$\uparrow$
& \textbf{Feas.}$\uparrow$ & \textbf{Score}$\uparrow$
& \textbf{Feas.}$\uparrow$ & \textbf{Score}$\uparrow$ \\
\midrule
Text-only
& 22.20\% & 0.0670
& 27.77\% & 0.0460
& 17.19\% & 0.0630 \\

EoH
& 87.80\% & 0.3052
& 21.00\% & 0.0860
& \underline{97.80\%} & \underline{0.2460} \\

FunSearch
& \underline{91.10\%} & \underline{0.3232}
& \underline{32.00\%} & \underline{0.1537}
& 81.10\% & 0.1456 \\

\midrule
Ours
& \textbf{95.69\%} & \textbf{0.3490}
& \textbf{100.00\%} & \textbf{0.4780}
& \textbf{98.15\%} & \textbf{0.4530} \\
\bottomrule
\end{tabular}
}
\end{table}

\subsection{Ablation Study}
\label{subsec:ablation}

We conduct ablation studies from two aspects: the sandbox-based planning scaffold and the atomic-task training strategy. The first study tests whether executable planning helps a frozen LLM, while the second examines the contribution of SFT, GRPO, ATRL, iterative deployment, and model scaling.

\paragraph{Effect of UrbanSandbox.}
We first evaluate four test-time strategies using DeepSeek as a frozen planner without post-training. \textbf{Text-only prompting} allows multi-turn reasoning but disables tool calls and evaluator execution. \textbf{EoH}~\cite{liuevolution} and \textbf{FunSearch}~\cite{FunSearch2023} are adapted as LLM-guided heuristic program search methods, where generated programs are evaluated by the task evaluator. \textbf{Ours} uses the UrbanSandbox-based atomic workflow, allowing the frozen LLM to inspect files, execute commands, evaluate plans, and revise outputs.
Table~\ref{tab:deepseek_scaffold} reports the results. Text-only prompting has feasibility below 30\% on all tasks, showing that direct generation is insufficient for constrained planning. EoH and FunSearch improve some results but remain unstable, especially on Road Planning, with only 21.00\% and 32.00\% feasibility. In contrast, our scaffold achieves the best feasibility and score on all tasks, confirming that executable file inspection, evaluator feedback, and structured revision substantially improve test-time planning reliability.

\paragraph{Effect of atomic-task training.}
We then evaluate training and deployment components under the full UrbanSandbox environment. All variants share the same setting unless otherwise specified. We compare \textbf{Qwen3-8B}, the raw model; \textbf{+SFT}, supervised fine-tuning on sandbox trajectories; \textbf{+GRPO}, flat RL over complete trajectories; \textbf{+ATRL}, atomic-task RL over BuildPlan and ImprovePlan; \textbf{CityPlanner}, which adds iterative refinement; and \textbf{+Qwen3-14B}, which replaces the 8B base model with Qwen3-14B under the same pipeline.
Table~\ref{tab:training_deployment_ablation} reports the results. The raw model has feasibility below 25\%, indicating that instruction following alone is insufficient. SFT raises feasibility above 80\%, while GRPO further improves objective scores. ATRL brings the largest gain, improving feasibility over GRPO by 8.3 percentage points on average, reaching 100.00\% feasibility on all tasks, and improving objective scores by 20.7\% on average. CityPlanner further improves objectives by 4.9\% over ATRL through iterative refinement, and Qwen3-14B adds another 2.8\% average gain. Overall, these results verify that each component contributes to feasibility and quality.

\begin{table}[t]
\centering
\caption{Ablation under the full UrbanSandbox environment. 
ATRL denotes atomic-task RL, and CityPlanner further applies iterative refinement deployment.}
\label{tab:training_deployment_ablation}
\resizebox{\columnwidth}{!}{
\begin{tabular}{lcccccc}
\toprule
\multirow{2}{*}{\textbf{Method}}
& \multicolumn{2}{c}{\textbf{Land Allocation}}
& \multicolumn{2}{c}{\textbf{Road Construction}}
& \multicolumn{2}{c}{\textbf{Station Placement}} \\
\cmidrule(lr){2-3}
\cmidrule(lr){4-5}
\cmidrule(lr){6-7}
& \textbf{Feas.}$\uparrow$ & \textbf{Obj.}$\uparrow$
& \textbf{Feas.}$\uparrow$ & \textbf{Obj.}$\uparrow$
& \textbf{Feas.}$\uparrow$ & \textbf{Obj.}$\uparrow$ \\
\midrule
Qwen3-8B
& 15.29\% & 0.1067
& 21.95\% & 0.1290
& 23.31\% & 0.1020 \\

\quad + SFT
& 82.40\% & 0.3120
& 87.10\% & 0.3077
& 82.40\% & 0.3120 \\

\quad + GRPO
& \underline{89.60\%} & 0.3440
& \underline{94.20\%} & 0.5860
& \underline{91.30\%} & 0.3790 \\

\quad + ATRL
& \textbf{100.00\%} & \underline{0.4210}
& \textbf{100.00\%} & \underline{0.6820}
& \textbf{100.00\%} & \underline{0.4670} \\

\midrule
CityPlanner
& \textbf{100.00\%} & \underline{0.4374}
& \textbf{100.00\%} & \underline{0.7163}
& \textbf{100.00\%} & \underline{0.4945} \\

\quad+ Qwen14B
& \textbf{100.00\%} & \textbf{0.4498}
& \textbf{100.00\%} & \textbf{0.7312}
& \textbf{100.00\%} & \textbf{0.5116} \\
\bottomrule
\end{tabular}
}
\end{table}

\begin{figure}[t]
    \centering
    \includegraphics[width=\linewidth]{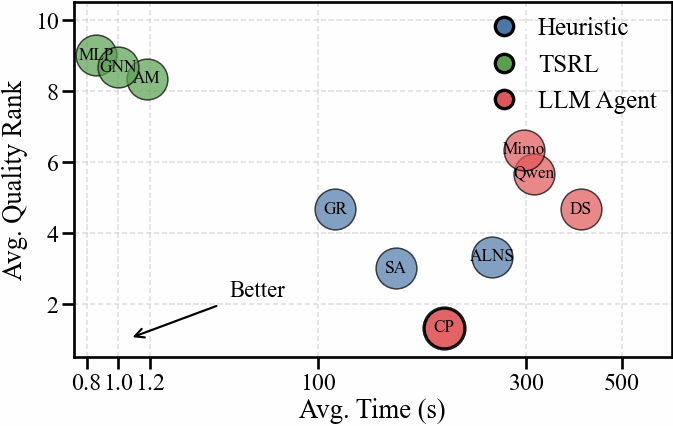}
    \label{fig:quality_time_bubble}
    \caption{Efficiency analysis of all methods. }
    \label{fig:efficiency_analysis}
\end{figure}

\subsection{Efficiency Analysis}
\label{subsec:efficiency}

We further evaluate CityPlanner from the efficiency perspective. Figure~\ref{fig:efficiency_analysis} compares different methods by their average planning quality and runtime. TSRL methods are the fastest because they only require lightweight policy inference, but their solution quality is limited. Heuristic methods achieve competitive quality on some tasks, but require non-trivial search time. General LLM agents are slower due to multi-turn reasoning and sandbox interaction. CityPlanner achieves the best overall planning quality while maintaining moderate runtime. This shows that the proposed executable planning workflow does not obtain higher quality by simply spending excessive search time.

\section{Conclusion}
\label{sec:conclusion}

In this paper, we presented CityPlanner, a sandbox-agent framework for executable urban planning. CityPlanner formulates heterogeneous planning tasks as sandbox interaction and introduces UrbanSandbox to provide a unified file-based environment with task-specific constraints, evaluators, and feedback. To make policy learning tractable, CityPlanner decomposes long and noisy planning trajectories into two atomic tasks, BuildPlan and ImprovePlan, and composes the learned skills through iterative refinement at inference time. Experiments on an OpenStreetMap-based benchmark show that CityPlanner consistently improves planning quality over heuristic, task-specific RL, and LLM-agent baselines. Further ablations verify the contributions of UrbanSandbox, atomic-task reinforcement learning, and iterative deployment. These results suggest that executable interaction is a promising interface for applying LLM agents to constrained urban decision-making.

\clearpage  
\section*{Limitations}

Despite its effectiveness, CityPlanner has several limitations. 
(1) \textbf{Computational cost.} CityPlanner requires multi-turn sandbox interaction, including file inspection, script execution, and repeated evaluator calls, making it slower than direct policy inference or lightweight heuristics. 
(2) \textbf{Optimality guarantee.} CityPlanner improves plans through iterative refinement, but it does not provide theoretical optimality guarantees. The final solution may still depend on model capability, reward design, and the informativeness of evaluator feedback.

\section*{Ethical Considerations}
\label{sec:ethics}

Our benchmark is built from OpenStreetMap, an open geographic data platform. We only use public map elements such as roads, land parcels, facilities, and points of interest, and do not use personal trajectories, demographic profiles, land-ownership records, or other sensitive user-level information. Thus, the dataset construction does not introduce additional privacy risks beyond the underlying public map data.

\bibliography{custom}

\begin{thebibliography}{35}
\providecommand{\natexlab}[1]{#1}

\bibitem[{mim(2026)}]{mimov25}
 2026.
\newblock Mimo-v2.5.
\newblock \url{https://huggingface.co/collections/XiaomiMiMo/mimo-v25}.

\bibitem[{Agrawal and Goktas(2025)}]{agrawal2025large}
Kushagra Agrawal and Polat Goktas. 2025.
\newblock How large language models transform urban planning and shape tomorrow’s cities?
\newblock In \emph{Large Language Models for Sustainable Urban Development}, pages 185--218. Springer.

\bibitem[{Ahuja et~al.(2000)Ahuja, Orlin, and Sharma}]{ahuja2000very}
Ravindra~K Ahuja, James~B Orlin, and Dushyant Sharma. 2000.
\newblock Very large-scale neighborhood search.
\newblock \emph{International Transactions in Operational Research}, 7(4-5):301--317.

\bibitem[{Cheng et~al.(2025)Cheng, Ouyang, Yu, Yan, Luo, Liu, Wang, Liu, and Chen}]{cheng2025agent}
Mingyue Cheng, Jie Ouyang, Shuo Yu, Ruiran Yan, Yucong Luo, Zirui Liu, Daoyu Wang, Qi~Liu, and Enhong Chen. 2025.
\newblock Agent-r1: Training powerful llm agents with end-to-end reinforcement learning.
\newblock \emph{arXiv preprint arXiv:2511.14460}.

\bibitem[{DeepSeek-AI(2026)}]{deepseekv4}
DeepSeek-AI. 2026.
\newblock Deepseek-v4: Towards highly efficient million-token context intelligence.

\bibitem[{Feng et~al.(2025)Feng, Huang, Qu, Zhang, Qin, Zhong, Jiang, Chi, and Zhong}]{feng2025retool}
Jiazhan Feng, Shijue Huang, Xingwei Qu, Ge~Zhang, Yujia Qin, Baoquan Zhong, Chengquan Jiang, Jinxin Chi, and Wanjun Zhong. 2025.
\newblock Retool: Reinforcement learning for strategic tool use in llms.
\newblock \emph{arXiv e-prints}, pages arXiv--2504.

\bibitem[{Feo and Resende(1995)}]{feo1995greedy}
Thomas~A Feo and Mauricio~GC Resende. 1995.
\newblock Greedy randomized adaptive search procedures.
\newblock \emph{Journal of global optimization}, 6(2):109--133.

\bibitem[{Jiang et~al.(2024)Jiang, Chao, Chen, Li, Liu, and Cong}]{jiang2024urbanllm}
Yue Jiang, Qin Chao, Yile Chen, Xiucheng Li, Shuai Liu, and Gao Cong. 2024.
\newblock Urbanllm: Autonomous urban activity planning and management with large language models.
\newblock In \emph{2024 Findings of the Association for Computational Linguistics, EMNLP 2024}, pages 1810--1825. Association for Computational Linguistics (ACL).

\bibitem[{Li et~al.(2025)Li, Hu, and Wang}]{li2025encouraging}
Zhiwei Li, Yong Hu, and Wenqing Wang. 2025.
\newblock Encouraging good processes without the need for good answers: Reinforcement learning for llm agent planning.
\newblock In \emph{Proceedings of the 2025 Conference on Empirical Methods in Natural Language Processing: Industry Track}, pages 1654--1666.

\bibitem[{Liu et~al.()Liu, Xialiang, Yuan, Lin, Luo, Wang, Lu, and Zhang}]{liuevolution}
Fei Liu, Tong Xialiang, Mingxuan Yuan, Xi~Lin, Fu~Luo, Zhenkun Wang, Zhichao Lu, and Qingfu Zhang.
\newblock Evolution of heuristics: Towards efficient automatic algorithm design using large language model.
\newblock In \emph{Forty-first International Conference on Machine Learning}.

\bibitem[{Liu et~al.(2023)Liu, Sun, and Qi}]{liu2023optimal}
Jiaqi Liu, Jian Sun, and Xiao Qi. 2023.
\newblock Optimal placement of charging stations in road networks: a reinforcement learning approach with attention mechanism.
\newblock \emph{Applied Sciences}, 13(14):8473.

\bibitem[{Liu et~al.(2025)Liu, Yigitcanlar, Mehmood, Corchado, and Fu}]{liu2025large}
Ke~Liu, Tan Yigitcanlar, Rashid Mehmood, Juan Corchado, and Xinyu Fu. 2025.
\newblock Large language models in urban planning: a systematic review and conceptual framework.
\newblock \emph{Journal of Urban Technology}, pages 1--44.

\bibitem[{Merrill et~al.(2026)Merrill, Shaw, Carlini, Li, Raj, Bercovich, Shi, Shin, Walshe, Buchanan et~al.}]{merrill2026terminal}
Mike~A Merrill, Alexander~G Shaw, Nicholas Carlini, Boxuan Li, Harsh Raj, Ivan Bercovich, Lin Shi, Jeong~Yeon Shin, Thomas Walshe, E~Kelly Buchanan, and 1 others. 2026.
\newblock Terminal-bench: Benchmarking agents on hard, realistic tasks in command line interfaces.
\newblock \emph{arXiv preprint arXiv:2601.11868}.

\bibitem[{{OpenStreetMap contributors}(2017)}]{OpenStreetMap}
{OpenStreetMap contributors}. 2017.
\newblock {Planet dump retrieved from https://planet.osm.org }.
\newblock \url{ https://www.openstreetmap.org }.

\bibitem[{Patil et~al.()Patil, Zhang, Wang, and Gonzalez}]{patilgorilla}
Shishir~G Patil, Tianjun Zhang, Xin Wang, and Joseph~E Gonzalez.
\newblock Gorilla: Large language model connected with massive apis.
\newblock In \emph{The Thirty-eighth Annual Conference on Neural Information Processing Systems}.

\bibitem[{Pysklo et~al.(2026)Pysklo, Zhuravel, and Watson}]{pysklo2026agent}
Hubert~M Pysklo, Artem Zhuravel, and Patrick~D Watson. 2026.
\newblock Agent-diff: Benchmarking llm agents on enterprise api tasks via code execution with state-diff-based evaluation.
\newblock \emph{arXiv preprint arXiv:2602.11224}.

\bibitem[{Qi et~al.()Qi, Liu, Iong, Lai, Sun, Sun, Yang, Yang, Yao, Xu et~al.}]{qiwebrl}
Zehan Qi, Xiao Liu, Iat~Long Iong, Hanyu Lai, Xueqiao Sun, Jiadai Sun, Xinyue Yang, Yu~Yang, Shuntian Yao, Wei Xu, and 1 others.
\newblock Webrl: Training llm web agents via self-evolving online curriculum reinforcement learning.
\newblock In \emph{The Thirteenth International Conference on Learning Representations}.

\bibitem[{Qian et~al.(2026)Qian, Acikgoz, He, Wang, Chen, Hakkani-Tur, Tur, and Ji}]{qian2026toolrl}
Cheng Qian, Emre~Can Acikgoz, Qi~He, Hongru Wang, Xiusi Chen, Dilek Hakkani-Tur, Gokhan Tur, and Heng Ji. 2026.
\newblock Toolrl: Reward is all tool learning needs.
\newblock \emph{Advances in Neural Information Processing Systems}, 38:105523--105553.

\bibitem[{Qin et~al.()Qin, Liang, Ye, Zhu, Yan, Lu, Lin, Cong, Tang, Qian et~al.}]{qintoolllm}
Yujia Qin, Shihao Liang, Yining Ye, Kunlun Zhu, Lan Yan, Yaxi Lu, Yankai Lin, Xin Cong, Xiangru Tang, Bill Qian, and 1 others.
\newblock Toolllm: Facilitating large language models to master 16000+ real-world apis.
\newblock In \emph{The Twelfth International Conference on Learning Representations}.

\bibitem[{Rasheed et~al.(2020)Rasheed, Yau, Noor, Wu, and Low}]{rasheed2020deep}
Faizan Rasheed, Kok-Lim~Alvin Yau, Rafidah~Md Noor, Celimuge Wu, and Yeh-Ching Low. 2020.
\newblock Deep reinforcement learning for traffic signal control: A review.
\newblock \emph{IEEE Access}, 8:208016--208044.

\bibitem[{Romera-Paredes et~al.(2023)Romera-Paredes, Barekatain, Novikov, Balog, Kumar, Dupont, Ruiz, Ellenberg, Wang, Fawzi, Kohli, and Fawzi}]{FunSearch2023}
Bernardino Romera-Paredes, Mohammadamin Barekatain, Alexander Novikov, Matej Balog, M.~Pawan Kumar, Emilien Dupont, Francisco J.~R. Ruiz, Jordan Ellenberg, Pengming Wang, Omar Fawzi, Pushmeet Kohli, and Alhussein Fawzi. 2023.
\newblock \href {https://doi.org/10.1038/s41586-023-06924-6} {Mathematical discoveries from program search with large language models}.
\newblock \emph{Nature}.

\bibitem[{Shao et~al.(2024)Shao, Wang, Zhu, Xu, Song, Bi, Zhang, Zhang, Li et~al.}]{grpo}
Zhihong Shao, Peiyi Wang, Qihao Zhu, Runxin Xu, Junxiao Song, Xiao Bi, Haowei Zhang, Mingchuan Zhang, YK~Li, and 1 others. 2024.
\newblock Deepseekmath: Pushing the limits of mathematical reasoning in open language models.
\newblock \emph{arXiv preprint arXiv:2402.03300}.

\bibitem[{Shinn et~al.(2023)Shinn, Cassano, Gopinath, Narasimhan, and Yao}]{shinn2023reflexion}
Noah Shinn, Federico Cassano, Ashwin Gopinath, Karthik Narasimhan, and Shunyu Yao. 2023.
\newblock Reflexion: Language agents with verbal reinforcement learning.
\newblock \emph{Advances in neural information processing systems}, 36:8634--8652.

\bibitem[{Tan et~al.(2025)Tan, Qu, Tu, Ge, Liu, Koehn, and Lu}]{tan2025process}
Weiting Tan, Xinghua Qu, Ming Tu, Meng Ge, Andy~T Liu, Philipp Koehn, and Lu~Lu. 2025.
\newblock Process-supervised reinforcement learning for interactive multimodal tool-use agents.
\newblock \emph{arXiv preprint arXiv:2509.14480}.

\bibitem[{Team(2025{\natexlab{a}})}]{openthoughts-agent}
OpenThoughts-Agent Team. 2025{\natexlab{a}}.
\newblock {OpenThoughts-Agent}.
\newblock https://www.open-thoughts.ai/blog/agent.

\bibitem[{Team(2025{\natexlab{b}})}]{ot-agent}
OpenThoughts-Agent Team. 2025{\natexlab{b}}.
\newblock {OpenThoughts-Agent}.
\newblock https://www.open-thoughts.ai/blog/agent.

\bibitem[{Team(2026)}]{qwen35}
Qwen Team. 2026.
\newblock \href {https://qwen.ai/blog?id=qwen3.5} {Qwen3.5: Accelerating productivity with native multimodal agents}.

\bibitem[{Van~Laarhoven and Aarts(1987)}]{van1987simulated}
Peter~JM Van~Laarhoven and Emile~HL Aarts. 1987.
\newblock Simulated annealing.
\newblock In \emph{Simulated annealing: Theory and applications}, pages 7--15. Springer.

\bibitem[{Vishwakarma et~al.(2025)Vishwakarma, Agarwal, Patil, Devaguptapu, and Chandran}]{vishwakarma2025can}
Harsh Vishwakarma, Ankush Agarwal, Ojas Patil, Chaitanya Devaguptapu, and Mahesh Chandran. 2025.
\newblock Can llms help you at work? a sandbox for evaluating llm agents in enterprise environments.
\newblock In \emph{Proceedings of the 2025 Conference on Empirical Methods in Natural Language Processing}, pages 9178--9212.

\bibitem[{von Wahl et~al.(2022)von Wahl, Tempelmeier, Sao, and Demidova}]{von2022reinforcement}
Leonie von Wahl, Nicolas Tempelmeier, Ashutosh Sao, and Elena Demidova. 2022.
\newblock Reinforcement learning-based placement of charging stations in urban road networks.
\newblock In \emph{Proceedings of the 28th ACM SIGKDD Conference on Knowledge Discovery and Data Mining}, pages 3992--4000.

\bibitem[{Yao et~al.(2023)Yao, Zhao, Yu, Du, Shafran, Narasimhan, and Cao}]{yao2023react}
Shunyu Yao, Jeffrey Zhao, Dian Yu, Nan Du, Izhak Shafran, Karthik Narasimhan, and Yuan Cao. 2023.
\newblock React: Synergizing reasoning and acting in language models.
\newblock In \emph{11th International Conference on Learning Representations, ICLR 2023}.

\bibitem[{Zheng et~al.(2023)Zheng, Su, Ding, Jin, and Li}]{zheng2023road}
Yu~Zheng, Hongyuan Su, Jingtao Ding, Depeng Jin, and Yong Li. 2023.
\newblock Road planning for slums via deep reinforcement learning.
\newblock In \emph{Proceedings of the 29th ACM SIGKDD Conference on Knowledge Discovery and Data Mining}, pages 5695--5706.

\bibitem[{Zheng et~al.(2025)Zheng, Xu, Lin, Santi, Ratti, Wang, and Li}]{zheng2025urban}
Yu~Zheng, Fengli Xu, Yuming Lin, Paolo Santi, Carlo Ratti, Qi~R Wang, and Yong Li. 2025.
\newblock Urban planning in the era of large language models.
\newblock \emph{Nature computational science}, 5(9):727--736.

\bibitem[{Zhou et~al.()Zhou, Lin, and Li}]{zhou2402large}
Z~Zhou, Y~Lin, and Y~Li.
\newblock Large language model empowered participatory urban planning. arxiv 2024.
\newblock \emph{arXiv preprint arXiv:2402.01698}.

\bibitem[{Zhu et~al.(2024)Zhu, Zhang, Huang, Li, Niu, Fan, Lun, Tao, Su, Gong et~al.}]{zhu2024plangpt}
He~Zhu, Wenjia Zhang, Nuoxian Huang, Boyang Li, Luyao Niu, Zipei Fan, Tianle Lun, Yicheng Tao, Junyou Su, Zhaoya Gong, and 1 others. 2024.
\newblock Plangpt: Enhancing urban planning with tailored language model and efficient retrieval.
\newblock \emph{arXiv preprint arXiv:2402.19273}.

\end{thebibliography}

\appendix

\label{sec:appendix}
\section{Additional Experimental Analysis}
\label{app:additional_experiments}

This section provides additional analyses of CityPlanner beyond the
main comparison. We examine training dynamics, context evolution under atomic tasks, interaction cost, and refinement behavior.

\subsection{Training Dynamics}
\label{app:training_dynamics}

\begin{figure}[ht]
    \centering
    \includegraphics[width=0.95\linewidth]{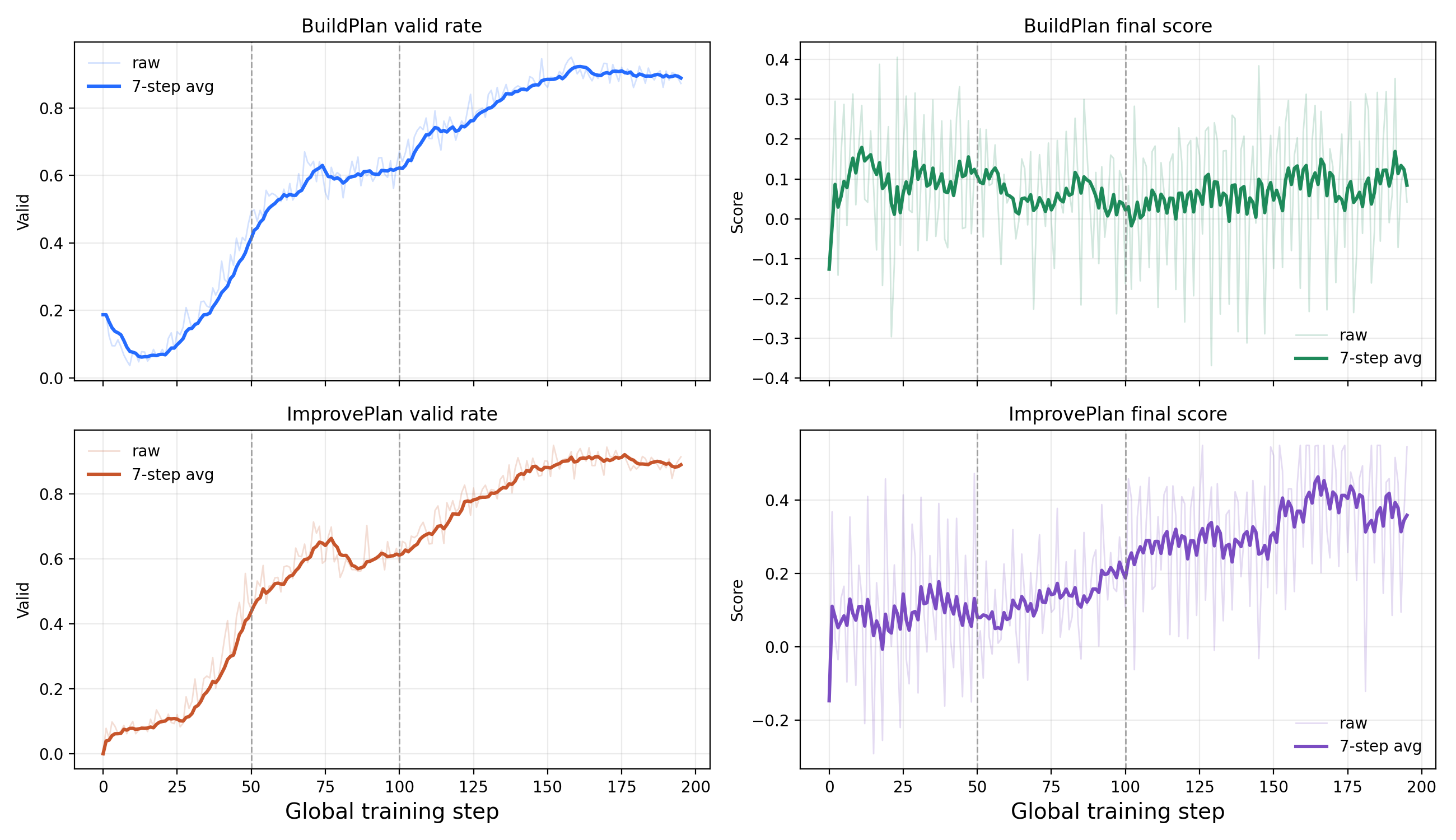}
    \caption{Training dynamics of atomic-task reinforcement learning.}
    \label{fig:training_dynamics}
\end{figure}

Figure~\ref{fig:training_dynamics} shows the training dynamics of
atomic-task reinforcement learning. The light curves show raw values,
and the bold curves show 7-step moving averages. For BuildPlan, the
valid rate increases steadily and eventually approaches a high success
rate, indicating that the model learns to construct feasible initial
plans. Its final score remains noisy, suggesting that the construction
task mainly improves validity rather than fine-grained quality
optimization. For ImprovePlan, both the valid rate and final score
increase over training. In particular, the score curve shows a clear
upward trend, indicating that the model learns to use evaluator
feedback to revise existing plans. These results show that the two
atomic tasks acquire complementary skills: BuildPlan stabilizes initial
feasible construction, while ImprovePlan improves plan quality through
feedback-based refinement.

\subsection{Context Evolution}
\label{app:context_evolution}

\begin{figure}[t]
    \centering
    \includegraphics[width=\linewidth]{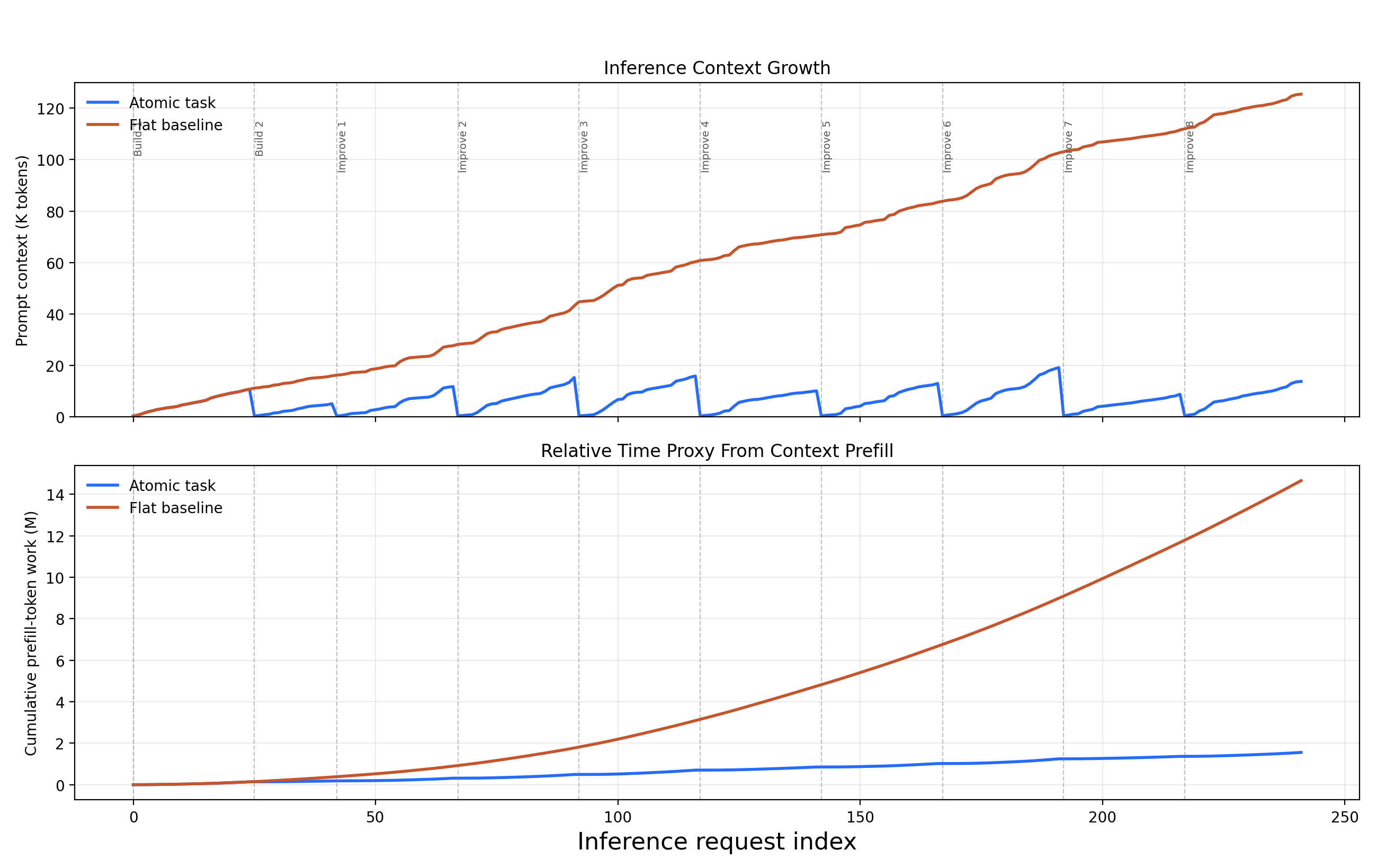}
    \caption{Context evolution under flat sandbox planning and atomic-task planning.}
    \label{fig:context_evolution}
\end{figure}

Figure~\ref{fig:context_evolution} compares the context growth of flat
sandbox planning and atomic-task planning during inference. In the flat
baseline, the prompt context grows monotonically as file contents,
command outputs, failed attempts, and intermediate revisions accumulate
in a single trajectory. In contrast, atomic-task planning resets the
context at each BuildPlan or ImprovePlan call, keeping each interaction
within a bounded context. This also leads to a much smaller cumulative
prefill-token workload, as shown in the lower panel. These results
support the motivation of atomic-task decomposition: it reduces context
accumulation while preserving the ability to solve long-horizon planning
through repeated bounded interactions. 

\subsection{Interaction Turns}
\label{app:interaction_turns}

\begin{figure}[t]
    \centering
    \includegraphics[width=\linewidth]{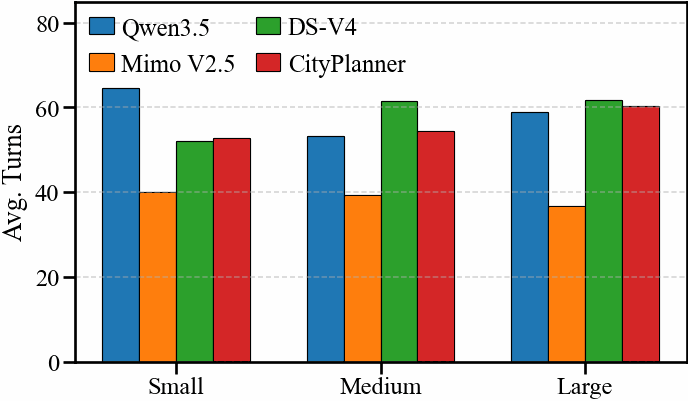}
    \caption{Average interaction turns of LLM agents under different task difficulties.}
    \label{fig:avg_turn}
\end{figure}

Figure~\ref{fig:avg_turn} reports the average number of interaction
turns used by LLM agents across difficulty levels. CityPlanner uses a
comparable number of turns to other strong LLM agents, indicating that
its performance gain does not come from substantially longer
interactions. Together with the main quality results, this suggests
that UrbanSandbox feedback and iterative refinement improve the
effectiveness of each interaction.

\subsection{Refinement Behavior}
\label{app:refinement_behavior}

\begin{figure}[t]
    \centering
    \includegraphics[width=0.9\linewidth]{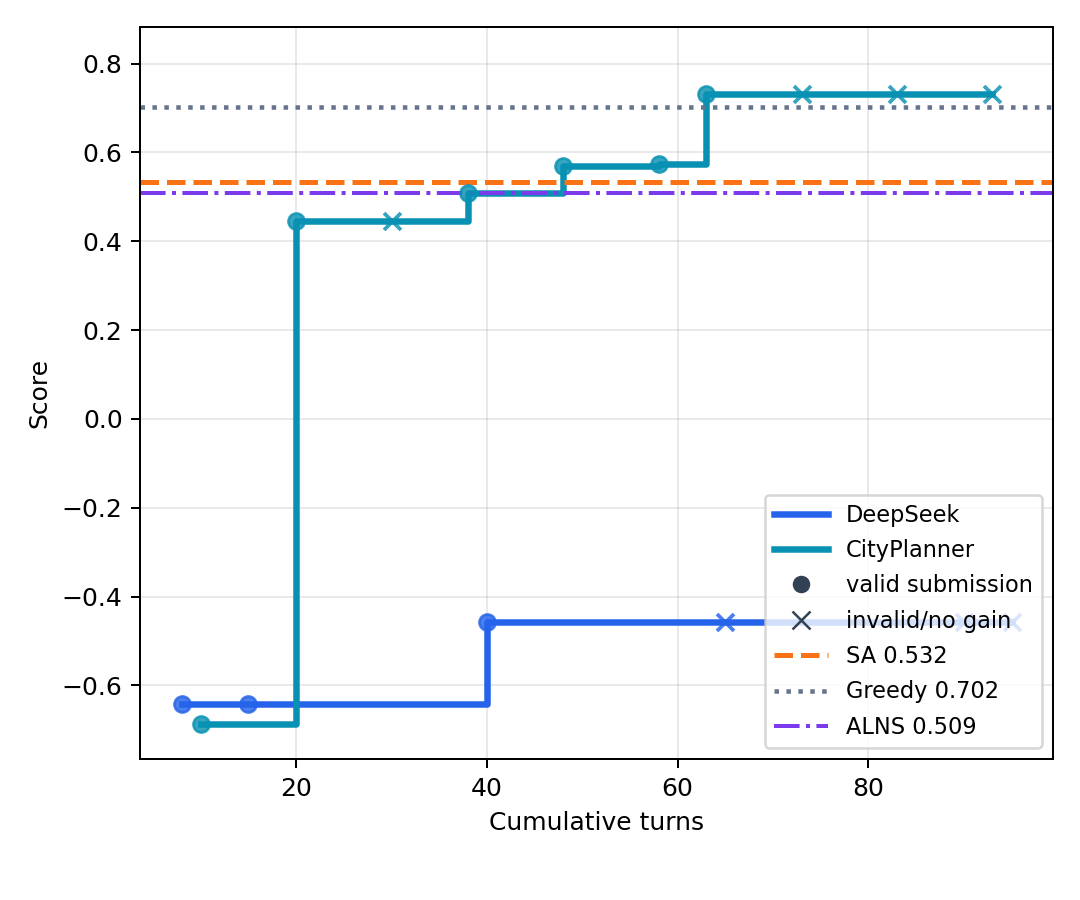}
    \caption{Best-so-far refinement curve on a representative road construction instance.}
    \label{fig:best_curve_road}
\end{figure}

Figure~\ref{fig:best_curve_road} shows the best-so-far score during a
representative road construction episode. DeepSeek remains in a
low-score region, indicating that direct long-horizon planning has
difficulty recovering from invalid or non-improving submissions. In
contrast, CityPlanner quickly obtains a valid intermediate plan and
then improves it through evaluator feedback. Its best-so-far score
first surpasses ALNS and SA, and eventually exceeds the Greedy baseline.
This result shows that iterative deployment enables stable plan
improvement by preserving the best-so-far solution instead of
regenerating plans from scratch.

\section{Task Definitions}
\label{app:task_definitions}

This section gives the task-specific definitions used in our benchmark.
All tasks are evaluated through a unified scoring interface. Submitted
action ids are first mapped to feasible candidates. Invalid ids are
recorded as violations, and duplicated action ids are removed before
scoring. Unless otherwise specified, all benchmark tables report the
raw evaluator scores defined below. Higher scores indicate better
planning quality for all tasks.

\subsection{Land Allocation}
\label{app:urban_definition}

\paragraph{Task origin.}
Land allocation is adapted from community spatial planning, where the
planner assigns land-use functions to parcels to improve spatial
efficiency under service, ecological, and mobility requirements
\citep{zheng2025urban}. In our benchmark, this task is instantiated as
selecting parcel--land-use actions that form a allocation plan.

\paragraph{Observation.}
An instance provides candidate parcels, demand
distributions, budget information, and a land-use requirement
configuration \texttt{need\_config}. The requirement configuration
specifies how many parcels of each land-use type are needed. Candidate
actions also contain cost, parcel attributes, and estimated service
effects.

\paragraph{Action.}
A plan $P$ selects a subset of candidate assignments. Each selected
action assigns one land-use type to one candidate parcel. A valid plan
should satisfy the required land-use composition, avoid duplicate block
selection, and respect budget and count constraints.

\paragraph{Score.}
The evaluator measures two main terms. The first is requirement
satisfaction $R_p(P)$, which measures how well selected land uses match
\texttt{need\_config}. The second is service coverage $A_p(P)$, which
measures the fraction of demand weight served by selected assignments.
The task-specific score is
\begin{equation}\small
S_p(P)
=
w_r R_p(P)
+
w_a A_p(P)
-
w_\rho \rho_p(P)
-
V_p(P),
\end{equation}
where $\rho_p(P)$ is the cost ratio and $V_p(P)$ is the violation
penalty. Following the original spatial-planning objective design, we
set $w_r=0.6$, $w_a=0.3$, and $w_\rho=0.2$. The violation term includes
duplicate select, overbuilding, and count violations.

\subsection{Road Construction}
\label{app:road_definition}

\paragraph{Task origin.}
Road construction is adapted from road planning for connectivity
improvement, where insufficient roads make urban places difficult to
access and limit the delivery of basic services \citep{zheng2023road}.
The original task selects new road segments from candidate locations
between places, with the goal of improving connectivity, reducing
travel distance, and controlling construct cost.

\paragraph{Observation.}
An instance provides an existing road network, demand groups, candidate
road segments, segment costs, and a construction budget. Each demand
group may contain multiple entry nodes. Candidate segments contain
endpoints, geometry, length, adjacent parcels, and construct cost.

\paragraph{Action.}
A plan $P$ selects a subset of candidate road segments to add to the
existing network. The evaluator constructs the planned road graph by
combining existing road edges and selected candidate edges. A valid plan
must satisfy budget and count constraints and should connect demand
groups through the resulting road graph.

\paragraph{Score.}
Let $\Gamma_p(P)$ denote the fraction of connected demand-group pairs.
If all demand pairs are connected, the evaluator computes a
demand distance under $P$ with that of an ideal graph containing all
feasible candidate roads. We define
\begin{equation}\small
Q_p(P)=
\begin{cases}
-1+\Gamma_p(P), & \Gamma_p(P)<1,\\
\Delta_p(P), & \Gamma_p(P)=1.
\end{cases}
\end{equation}
The task-specific score is
\begin{equation}\small
S_p(P)
=
Q_p(P)
-
w_\rho \rho_p(P)
-
V_p(P),
\end{equation}
where $\rho_p(P)$ is the construct-cost ratio and $V_p(P)$ is the
violation penalty. We set $w_\rho=0.1$ following the road-planning
objective. Thus, connectivity is the primary criterion, and fully
connected plans are further ranked by distance reduce and cost.

\subsection{Station Placement}
\label{app:ev_definition}

\paragraph{Task origin.}
Station placement is adapted from charging-station placement in urban
road networks \citep{von2022reinforcement}. The original task decides
where to place charging stations and what charger configuration each
site should use, with the goal of improving charging supply while
reducing access distance, service time, and construction cost.

\paragraph{Observation.}
An instance provides a road network, charging demand, candidate
charging sites, charger configs, and cost constraints. Each
candidate contains a site location and the numbers of slow, medium, and
fast chargers. The evaluator also uses demand weights, access
distances, capacity, and load information to estimate service quality.

\paragraph{Action.}
A plan $P$ selects candidate site--configuration actions. Each selected
action places a charger configuration at one site. A valid plan should
avoid selecting multiple configurations at the same site, satisfy count
and budget constraints, and provide sufficient capacity for nearby
demand.

\paragraph{Score.}
For each selected site $s$, the evaluator computes an effective supply
from the charger config:
\begin{equation}\small
q_p(s)
=
n_{\mathrm{slow}}(s)
+
3n_{\mathrm{medium}}(s)
+
8n_{\mathrm{fast}}(s).
\end{equation}
Demand is assigned to selected sites using a generalized distance that
combines physical distance and queue pressure. The evaluator then
computes service coverage $A_p(P)$, access-distance $D_p(P)$,
and service-time $T_p(P)$. The benefit term is
\begin{equation}\small
B_p(P)
=
w_a A_p(P)
+
w_d D_p(P)
+
w_t T_p(P).
\end{equation}
The task-specific score is
\begin{equation}\small
S_p(P)
=
B_p(P)
-
w_\rho \rho_p(P)
-
V_p(P),
\end{equation}
where $\rho_p(P)$ is the cost ratio and $V_p(P)$ aggregates duplicate
site, count, overload, and unserved-demand penalties. Following the
charging-station placement objective, we set $w_a=0.50$, $w_d=0.25$,
$w_t=0.15$, and $w_\rho=0.20$.

\subsection{Summary}
\label{app:score_summary}

The three tasks share the same evaluator convention: a submitted plan
is mapped to feasible candidates, checked against task-specific
constraints, and assigned a raw objective score $S_p(P)$. The main
experiments report these raw scores directly. Reward normalization,
if used during training, is not part of the benchmark metric.

\section{Dataset Construction}
\label{app:dataset_details}

\begin{table}[t]
\centering
\caption{UrbanSandbox dataset statistics across five urban clusters.}
\label{tab:urban_clusters}
\resizebox{\columnwidth}{!}{
\begin{tabular}{lcc}
\toprule
\textbf{Region} & \textbf{Tiles} & \textbf{Representative Cities} \\
\midrule
Beijing--Tianjin--Hebei & 25,269 & Beijing, Tianjin, Shijiazhuang \\
Yangtze River Delta & 21,872 & Shanghai, Nanjing, Hangzhou \\
Chengdu--Chongqing & 19,266 & Chengdu, Chongqing, Mianyang \\
Mid--Yangtze & 19,488 & Wuhan, Yichang, Changsha \\
Pearl River Delta & 17,203 & Guangzhou, Shenzhen, Hong Kong \\
\bottomrule
\end{tabular}
}
\end{table}

This section describes the construction of our OpenStreetMap-based benchmark. Each urban tile covers a geographic region of $0.03^\circ \times 0.03^\circ$, corresponding to approximately $3.3$ km by $2.6$ km depending on latitude. We first collect large-scale OpenStreetMap road-network data from five major urban clusters in China, covering 15 representative cities. As summarized in Table~\ref{tab:urban_clusters}, the initial crawl contains 103,098 tiles across Beijing--Tianjin--Hebei, Yangtze River Delta, Chengdu--Chongqing, Mid--Yangtze, and Pearl River Delta.

We then polygonize the road topology to identify urban faces enclosed by roads, which are used as spatial units for candidate generation and demand modeling. To ensure reliable instance construction, we retain only highly urbanized tiles with dense road networks and sufficient residential or commercial land-use evidence. After filtering low-density or unreliable regions, we obtain 1,890 final urban tiles. The tiles are split into 1,512 training tiles and 378 test tiles, following an 80/20 split.

Each tile is used to instantiate three planning tasks: land allocation, road construction, and station placement. Therefore, the final benchmark contains
5,670 planning instances. The three tasks share the same geographic region within each tile, but differ in candidate generation, constraints, and evaluator definitions.

\paragraph{Difficulty levels.}
We assign each instance to a size bucket according to the number of candidate actions and demand regions. The benchmark contains three difficulty levels: small, medium, and large. On average, road construction contains about 63, 102, and 143 candidates for small, medium, and large instances, respectively. Station placement contains about 80, 80, and 180 candidates, while land allocation contains about 85, 119, and 199 candidates. Thus, larger instances correspond to larger candidate spaces and more complex constraints.

\paragraph{City graph fields.}
Each instance contains a structured city state. Road-construction instances include junction nodes with geographic coordinates, junction degree, and adjacent road ids. All tasks include road edges with geometry, length, center coordinates, OSM road type, and source information. Demand zones contain parcel ids, center coordinates, area, demand weight, and task-specific entry nodes or geometry when applicable. These fields provide the spatial and attribute information used by UrbanSandbox evaluators.

\paragraph{Candidate actions.}
Each task provides a list of feasible candidate actions. For station placement, a candidate specifies a charging site and charger configuration, including the numbers of slow, medium, and fast chargers. For road construction, a candidate specifies a road segment with endpoints, geometry, length, cost, and adjacent parcels. For land allocation, a candidate assigns a land-use type to a block and includes its area, cost, and estimated service effects. All candidate actions are associated with an \texttt{action\_id}, cost, feasibility flag, payload, and estimated effects.

\section{Sandbox Implementation}
\label{app:sandbox_prompt}

This section describes the concrete workspace interface and prompt
protocol used by UrbanSandbox. Each episode creates a workspace with
four directories: \texttt{input/}, \texttt{work/}, \texttt{outputs/},
and \texttt{logs/}. The \texttt{input/} directory is read-only and
contains task data and evaluator scripts. The \texttt{work/} directory
is used for scratch files and intermediate analysis. The
\texttt{outputs/} directory stores submitted plans and evaluation
results, while \texttt{logs/} records traces.

\begin{interfacebox}{Workspace Interface}
\texttt{input/}: read-only task files, candidate files, city state, and evaluator scripts. \\
\texttt{work/}: writable scratch space and atomic-task context. \\
\texttt{outputs/}: submitted plans and evaluation history. \\
\texttt{logs/}: execution traces.
\end{interfacebox}

At initialization, \texttt{input/} contains the visible city state,
candidate schema, compact candidate tables, the full candidate list,
the complete task export, and evaluator scripts. In particular,
\texttt{city\_state.json} records the task name, instance id, metadata,
budget, interaction limit, candidate count, and demand count.
\texttt{candidate\_summary.csv} and \texttt{candidates.csv} provide
compact candidate views, while \texttt{candidates.json} stores the
complete candidate list. The file \texttt{export.json} contains the
complete city-task export, including demand regions, candidate actions,
road edges, nodes when applicable, and distance information.

Each atomic-task episode also provides
\texttt{work/atomic\_context.json}. For BuildPlan, this file contains
an empty current plan, the budget, the step limit, and editable
candidate information. For ImprovePlan, it additionally contains the
base plan and its score. Thus, the prompt itself does not contain
instance-specific city data. The model must retrieve the city graph,
candidates, constraints, current plan, and diagnostics by inspecting
workspace files and executing commands.

\begin{interfacebox}{Submitted Plan Format}
The submitted plan must be written to
\texttt{final\_plan.json} in the following format:
\[
\texttt{\{"candidate\_ids": ["id\_1", "id\_2"]\}}
\]
The evaluator accepts several compatible formats, but all prompts use
\texttt{candidate\_ids} as the official schema.
\end{interfacebox}

The agent evaluates a plan by running the evaluator command shown
below. The evaluator returns a JSON object containing validity,
validation diagnostics, task-specific score fields, and the parsed
candidate ids. Each evaluation is appended to
\texttt{outputs/eval\_history.jsonl}.


\paragraph{Terminal interaction protocol.}
All agents follow the same terminal protocol. At each interaction step,
the model must output one valid JSON object with four fields:
\texttt{analysis}, \texttt{plan}, \texttt{commands}, and
\texttt{task\_complete}. The \texttt{commands} field stores terminal
commands to execute, and \texttt{task\_complete} is set to
\texttt{true} only after the final plan has been written and evaluated.

\begin{promptlisting}{Example Terminal Response}
{
  "analysis": "I will inspect the task description first.",
  "plan": "Read the README and candidate summary.",
  "commands": [
    {
      "keystrokes": "cat input/README.md\n",
      "duration": 0.1
    }
  ],
  "task_complete": false
}
\end{promptlisting}

\subsection{Prompt Templates}
\label{app:prompts}

\begin{promptlisting}{System Prompt}
You are a helpful assistant that can interact with a computer to solve tasks.

You have access to a terminal. At each step you must respond with a single valid JSON object, and no other text.

Required JSON fields:
- analysis: brief analysis of the current terminal output and task progress
- plan: what you plan to do next
- commands: terminal commands to execute
- task_complete: whether the task is finished

Rules:
- Every command string in "keystrokes" must end with "\n"
- Multiple commands may be included and will execute sequentially
- Set "task_complete" to true only after verifying the task is complete
- Duration is 0.1 for instant commands, 1.0 for normal commands, and at most 60
- The response must be only the JSON object
\end{promptlisting}

\begin{promptlisting}{Instance Prompt Template}
Please solve this city-planning workspace task: {{task}}

Instance Context:
- instance_id: {{instance_id}}
- task: {{task_name}}
- size_bucket: {{size_bucket}}
- split: {{split}}

Workspace I/O Contract:
You are already at the workspace root.
Read task data from input/. Start with input/README.md and inspect candidate files with compact Python summaries.
Use work/ only for scratch scripts and intermediate files.
The only submitted plan is outputs/final_plan.json.

Goal:
{{ATOMIC_GOAL}}

Plan format:
{"candidate_ids": ["candidate_id_1", "candidate_id_2"]}

Evaluate the submitted plan before finishing:
python input/evaluate_plan.py outputs/final_plan.json

The evaluator returns validity, score, task-specific metrics, cost ratio, and hard errors. Repair hard errors before finishing.

When done, set "task_complete": true.

Workspace State:
- plan: exists={{plan_exists}} | non_empty={{plan_non_empty}} | selected_count={{selected_count}} | needs_eval={{needs_eval}}
- evaluation: has_evaluated={{has_evaluated}} | valid={{last_valid}} | score={{last_score}}
- hard_errors: {{hard_errors}}
- warnings: {{warnings}}
- next: {{next_required_action}}
\end{promptlisting}

BuildPlan and ImprovePlan share the same system prompt and workspace
I/O contract. The prompt specifies the terminal response protocol,
workspace files, final-plan schema, evaluator command, and completion
rule. The only task-specific part is the atomic goal. Importantly, the
prompt does not contain the full city graph, candidate table, demand
regions, distance, or geometry; these data must be retrieved
from the workspace files.

\begin{compactbox}{Atomic Goals}
\textbf{BuildPlan.}
Create a valid and non-empty initial plan for the current workspace.
The plan need not be globally optimal, but it should be executable and evaluated.

\medskip
\textbf{ImprovePlan.}
Improve the preloaded plan for the current workspace. The model should
make and evaluate a concrete edit to \texttt{candidate\_ids} that aims
to increase the score while keeping the plan valid and non-empty.
\end{compactbox}

\section{Implementation Details}
\label{app:implementation}

This section summarizes the implementation details of CityPlanner and
baselines. All methods are evaluated with the same official task
evaluator. Submitted action ids are first mapped to feasible candidates and budget or maximum-step violations are treated as hard
invalid cases.

\paragraph{CityPlanner and LLM agents.}
CityPlanner uses Qwen3-8B as the base model unless otherwise specified.
All LLM agents follow the same terminal interaction protocol and submit
plans to \texttt{outputs/final\_plan.json} with the
\texttt{candidate\_ids} schema. During evaluation, we use temperature
0.7, a maximum response length of 4096 tokens, and a request timeout of
300 seconds. The evaluation script allows at most 20 turns, and uses 16
workers. In full CityPlanner inference, BuildPlan is retried up to three
times if needed, and ImprovePlan is repeated until a patience budget of
three non-improving rounds is reached.

\paragraph{Atomic-task RL.}
After SFT, we optimize the model with GRPO over BuildPlan and
ImprovePlan. The rollout batch size and validation batch size are both
64. We use learning rate \(1.0\times10^{-6}\), per-device batch size 1,
gradient accumulation steps 32, and rollout temperature 1.0. The actor
inference batch size is 1, and the environment inference batch size is
4.

\paragraph{Heuristic baselines.}
All heuristic baselines use the same raw evaluator score as CityPlanner.
The shared heuristic script uses 32 workers, seed 42, and a repair
candidate limit of 300. Greedy starts from an empty plan and repeatedly
adds the feasible candidate that maximizes score improvement until no
addition improves the score. SA runs for 2000 iterations with add,
remove, and swap moves. ALNS runs for 1000 iterations with random,
expensive, and worst-contribution destroy operators, and greedy or
noisy-greedy repair operators.

\paragraph{Task-specific RL baselines.}
We implement MLP+PPO, AM+PPO, and GNN+PPO with MaskablePPO. The default
training budget is 200K timesteps. We use 8 parallel environments,
learning rate \(3\times10^{-4}\), rollout length 128, batch size 64,
10 PPO epochs, discount factor 0.99, GAE parameter 0.95, clipping range
0.2, and entropy coefficient 0.01. The MLP policy uses hidden sizes
\([256,256]\). The attention and GNN policies use embedding dimension
64 with two layers and four heads.

\paragraph{Scaffold baselines.}
\textbf{EoH} and \textbf{FunSearch} are adapted as
LLM-guided heuristic program search methods. EoH follows an
evolutionary heuristic design process: the LLM proposes heuristic
programs for selecting candidate actions, the evaluator scores the
resulting plans, and high-performing programs are used to guide later
generations. FunSearch follows a similar program-search paradigm, where
the LLM iteratively generates scoring or selection functions that are
executed to produce candidate plans and ranked by evaluator feedback.
Both baselines therefore use the LLM to search over executable
heuristics rather than directly interacting with the full workspace as
an agent. Our scaffold-only variant instead uses the UrbanSandbox
atomic workflow with file inspection, command execution, evaluator
calls, and plan revision, but does not update model parameters.


\end{document}